\documentclass{article}

\usepackage{microtype}
\usepackage{graphicx}
\usepackage{subfigure}
\usepackage{booktabs} 
\usepackage{hyperref}

\usepackage[accepted]{icml2023}

\usepackage{amsmath}
\usepackage{amssymb}
\usepackage{mathtools}
\usepackage{amsthm}

\usepackage[capitalize,noabbrev]{cleveref}

\theoremstyle{plain}

\theoremstyle{definition}

\theoremstyle{remark}

\usepackage[textsize=tiny]{todonotes}

\usepackage{balance}
\usepackage{placeins}

\newcommand{\papertitle}{Estimating label quality and errors in semantic segmentation data via any  model}

\icmltitlerunning{\papertitle}

\begin{document}

\twocolumn[
\icmltitle{\papertitle}



\icmlsetsymbol{equal}{*}

\begin{icmlauthorlist}
\icmlauthor{Vedang Lad}{comp,sch}
\icmlauthor{Jonas Mueller}{comp}
\end{icmlauthorlist}

\icmlaffiliation{comp}{Cleanlab}
\icmlaffiliation{sch}{MIT}

\icmlcorrespondingauthor{Vedang Lad}{vedang@mit.edu}
\icmlcorrespondingauthor{Jonas Mueller}{jonas@cleanlab.ai}

\icmlkeywords{Machine Learning, ICML}

\vskip 0.3in
]

\printAffiliationsAndNotice{}  

\begin{abstract}
 The labor-intensive annotation process of semantic segmentation datasets is often prone to errors, since humans struggle to label every pixel correctly. 
 We study algorithms to automatically detect such annotation errors, in particular methods to score \emph{label quality}, such that the images with the lowest scores are least likely to be correctly labeled. This helps prioritize what data to review in order to ensure a high-quality training/evaluation dataset, which is critical in sensitive applications such as medical imaging and autonomous vehicles. Widely applicable, our label quality scores rely on probabilistic predictions from a trained segmentation model -- \emph{any} model architecture and training procedure can be utilized. Here we study 7 different label quality scoring methods used in conjunction with a DeepLabV3+ or a FPN segmentation model to detect annotation errors in a version of the SYNTHIA dataset. 
Precision-recall evaluations reveal a score -- the soft-minimum of the model-estimated likelihoods of each pixel's annotated class -- that is particularly effective to identify images that are mislabeled, across multiple types of annotation error. 
\end{abstract}

\section{Introduction}
\label{intro}

Semantic segmentation, in which a model classifies each pixel within an image, is a cornerstone task of fine-grained image understanding \cite{segmentanything}. 
To train more effective segmentation models, the size of image datasets has dramatically grown in recent years, especially in fields like radiology, pathology, robotics, and autonomous vehicles. 
The labeling of semantic segmentation data necessitates pixel-wise annotation of images, which is highly label-intensive and error-prone, thus often failing to produce the ``ground truth'' it is claimed to be. Training models to output incorrect labels is obviously problematic, but even evaluating models with noisy labels is worrisome in high-stakes medical/robotics applications built around the segmentation task. 

Even significantly simpler to annotate data, e.g.\ for multi-class/multi-label classification, contain many errors  \cite{northcutt2021pervasive, thyagarajan2022identifying}. 
To address this, there is a growing interest in data-centric algorithms, specifically Label Error Detection (LED) methods, which aim to systematically enhance the quality of these datasets \cite{kuan, northcutt2021confident, Muller_2019}. 
Here we consider LED in semantic segmentation data, where the prevalence of label errors is likely much greater than in classification data, due to the extra annotation complexity. We consider algorithms to assign a \emph{label quality score} to each image\footnote{Code to run our method: \\ \url{https://github.com/cleanlab/cleanlab}}, such that the images with the lowest scores are most likely to have some sort of annotation error. These are the images that should be prioritized for review or re-labeling \cite{eric, kuan, goh2023activelab}. An effective score will ensure reviewers do not wastefully inspect correctly-labeled images (high precision) while simultaneously overlooking few mislabeled images (high recall). 

In this work, we focus on universal LED methods that can be applied to any segmentation dataset for which someone has trained \emph{any} standard type of segmentation model. The methods considered here only require predictions from the model and the annotated labels for the images in the dataset, we do not require running inference on additional (synthesized) images \cite{rottmann2022automated, nishi2021augmentation}, nor training models in a special nonstandard way \citep{medicalmicrosoft,sukhbaatar2014learning,jiang2018mentornet,zhang2018generalized}. This ensures our methods are very easy to apply and widely applicable across domains. More importantly, the best segmentation architectures are constantly evolving, and LED methods that require nonstandard training/inference may no longer be compatible with future state-of-the-art 
architectures. Since the performance of the segmentation model is obviously important for using it to detect mislabeling, LED methods must remain compatible with the best architectures to provide the best insights about a dataset.

Our experiments study three common types of potential annotation error in a segmentation dataset, where for a specific image, annotators may: (1) overlook an object depicted in the image, i.e.\ its true segmentation mask has been dropped in the given annotation (\emph{Drop}), (2) select the incorrect class label to apply for an annotated segmentation mask, i.e.\ its true class has been swapped with another class in the given annotation (\emph{Swap}), or (3) miss some pixels which should have been in/out of the mask when annotating it, i.e.\ the mask's true proper location has been shifted in the given annotation (\emph{Shift}). 
While ``ground truth'' is often used to refer to the given annotation in the segmentation literature, throughout this paper, we sparingly use ``truth'' to refer to ideal segmentation masks that a hypothetical perfect annotator would produce. The labels we train a model to predict are merely \emph{noisy}  reflections of this underlying truth in most real-world datasets \cite{northcutt2021pervasive}.
Here we present a model-agnostic \emph{softmin} label quality score that is able to accurately detect all 3 types of annotation error, in order to help us avoid training models with untrue data.


\section{Related Work}
\label{Related Works}

A large body of work has studied robust ML with noisy labels \cite{song, wei2022learning, chen2019understanding, nata}, but the importance of \emph{label error detection} alone was only more recently recognized \cite{Muller_2019,northcutt2021pervasive,northcutt2021confident,kuan}. With growing value from algorithms to improve data quality, LED methods have been proposed for more complex supervised learning tasks including multi-label classification \cite{thyagarajan2022identifying}, sequence prediction in NLP \cite{eric,klie2022annotation}, and object detection \cite{chachula2022combating, murrugarra2022can}. \looseness=-1

Specifically for semantic segmentation, \citet{vuadineanu2022analysis} demonstrate the destructive effects of annotation errors in medical imaging data. 
For segmentation in medical imaging, \citet{medicalmicrosoft} introduce a supervised segmentation method for jointly estimating the spatial characteristics of label errors from multiple human annotators. However their approach is model-specific requiring two coupled CNNs trained in a special manner unlike the general label quality scores we consider here. 
\citet{medicalmicrosoft} also introduce the notion of \textit{under-segmentation}, \textit{over-segmentation}, and \emph{class swaps} as common types of errors in segmentation data, which we also study here. 

More similar to our work is the general principle of \emph{error analysis}, in which one sorts images by loss (or other evaluation metric) incurred between the model prediction and annotated label. The worst-predicted images often reveal suspicious annotations or other data anomalies. For segmentation one traditionally uses the likelihood loss or IoU evaluation metric. Both are considered as label quality scores in our study, but each has shortcomings. For LED in segmentation data, \citet{rottmann2022automated} found operating on connected components of each image to be more effective than pixel-level analysis, but such approaches have vastly worse computational complexity than the straightforward softmin label quality score we propose here.

\section{Methods for Scoring Label Quality}

Semantic segmentation datasets typically consist of many images, each containing a fixed grid of pixels. To label the data, annotators assign each pixel to one of $K$ possible classes (i.e.\ categories in the segmentation task). 
For a given image $x$, a standard semantic segmentation model $M$ generates predicted class probabilities $p = M(x)$, where $p_{ijk}$ represents the estimated  probability that the $i,j$-th pixel in image $x$ belongs to class $k$. 
For unbiased evaluation of label quality based on model predictions, we must avoid overfitting. Thus we assume throughout this work that these predicted probabilities are computed \emph{out-of-sample}, derived from a copy of the model that did not encounter $x$ during training. Using any type of model, out-of-sample predictions can be obtained for 
\emph{every} image in a dataset via say 5-fold cross-validation.

Utilizing $p$, we initially evaluate the individual fine-grained labels for each pixel. In this context, we adopt effective LED methodologies for standard classification scenarios \cite{northcutt2021confident, kuan}, treating every pixel as a distinct, independent instance (regardless of the image to which it belongs).

For one image with pixel dimensions $h \times w$, we thus have:
\begin{itemize}
\item $\mathbf{p} \in \mathbb{R}^{h \times w \times K}$, where $p_{ijk}$ is the model predicted probability that pixel $(i, j)$ belongs to class $k$.
\item $\mathbf{P}=\{P_{ij}\}$ is the predicted segmentation mask, where $P_{ij} \in \{0,...,K-1\}$ is the predicted class label of pixel $(i, j)$, i.e.\ the class with the highest probability $ = \underset{k}{\mathrm{argmax}}\: p_{ijk}$.
\item $\mathbf{l}=\{l_{ij}\}$ is the annotated segmentation mask, where $l_{ij} \in \{0,...,K-1\}$ is the annotated class label for pixel $(i, j)$.
\item $\mathbf{s}=\{s_{ij}\}$ is an numeric array of per-pixel scores, where $s_{ij}$ is a fine-grained label quality score for pixel $(i, j)$. Here we consider fine-grained scores computed via some simple function $s_{ij} = f(l_{ij}, p_{ij1},\dots,p_{ijK})$ applied independently to the information at each pixel $(i, j)$. Any label quality score for classification data could be used here \cite{kuan}. We stick with the most straightforward choice: $s_{ij} = p_{ijk^*}$ where $k^* = l_{ij}$, i.e. the model-estimated likelihood of the annotated class  \cite{Muller_2019}. 
\item $\mathbf{b}=\{b_{ij}\}$ is an array of per-pixel binary values, where $b_{ij}=0$ if the pixel at position $(i, j)$ is flagged as potentially mislabeled (otherwise $b_{ij}=1$) by a \emph{flagger} LED algorithm \cite{klie2022annotation}. Here we use  Confident Learning \cite{northcutt2021confident}. Such methods can be applied to estimate \emph{which} instances are mislabeled in a classification dataset; here we can simply treat each pixel as a separate instance for a direct application.
\end{itemize}

Having computed (some of) the above per-pixel information, we are interested in obtaining an overall label quality score $s$ for each image $x$, to help us prioritize which images require review.
To avoid wasted reviewing effort, our primary desiderata is that images with the \emph{lowest} scores should contain some form of annotation error, and images where every pixel is labeled correctly should receive much \emph{higher} scores. An effective score $s$ should be robust to minor statistical fluctuations in per-pixel model outputs which are inevitable in practical deep learning. 
In this paper, we consider the following 8 methods for producing the label quality score $s(x)$ of an image $x$. We first present some baseline approaches that help better motivate our proposed \emph{Softmin} method.

\subsection{Correctly Classified Pixels (CCP)}
This approach measures the proportion of correctly classified pixels in an image. It is calculated as follows:
\begin{equation}
s_{CCP}(x) = \frac{\sum_{i,j}\mathbb{I}[l_{ij}=P_{ij}]}{h \cdot w}
\end{equation}
Relying on \emph{Correctly Classified Pixels} is an intuitive way to use machine learning to verify the choices made by data annotators. 
This approach should be robust to small statistical fluctuations in the predicted probability values output by a model, given it only operates on the predicted segmentation mask. However it ignores both model confidence and shortcomings of the trained model.

\subsection{Thresholded CCP (TCCP)}
The Correctly Classified Pixels method is sensitive to the model's propensity to over/under predict certain classes relative to others. 
This alternative approach extends Correctly Classified Pixels with an adaptive threshold parameter and calculating accuracy using a different threshold for each class. We select the threshold value for each class that maximizes accuracy for predicting this class, and use the mean over all classes as our overall label quality score.

For each class $k \in \{1, \ldots, K\}$ and threshold $\tau$ in a predefined set of thresholds $\mathcal{T}$, we calculate accuracy as follows:

\begin{equation}
s_{TCCP, t}^{k}(x) = \frac{\sum_{i,j}\mathbb{I}[l_{ij}=k, p_{ijk}>\tau]}{h \cdot w}
\end{equation}
For each class $k$, we select a separate threshold value $\tau^{*}_k$ that maximizes $s_{TCCP, \tau}^{k}(x)$:
\begin{equation}
\tau^{*}_k = \underset{\tau \in T}{\mathrm{argmax}} \: s_{TCCP, \tau}^{k}(x)
\end{equation}

The overall TCCP label quality score for an image $x$ is then the mean of these per-class values over all classes:
\begin{equation}
s_{TCCP}(x) = \frac{1}{K} \sum_{k=1}^{K} s_{TCCP, \tau^{*}_k}^{k}(x)
\end{equation}
Thresholding the probabilistic predictions separately for each class allows us to account for shortcomings of the model if it systematically over/under-predicts a certain class. This can be viewed as a form of post-hoc calibration applied before computing the overall label quality score. 
\subsection{Confidence In Label (CIL)}
We compute the confidence of a given label using the model's predicted probabilities -- simply adopting the model-estimated likelihood of the annotated class at each pixel as a label quality score. 
Recall that $p_{i,j,c}$ denotes the predicted probability that pixel at position $(i, j)$ belongs to class $c$, and consider the shorthand:  
$s_{ij} := p_{i,j,l_{ij}}$, for expressing the model-estimated likelihood of the annotated class at each pixel $l_{ij}$. Each $s_{ij}$ serves as a quality score for an individual pixel annotation, and the CIL score of each image $x$ is simply the mean of these pixel-scores within the image:
\begin{equation}
\label{CIL}
s_{CIL}(x) = \frac{1}{h \cdot w} \sum_{i,j} s_{ij}
\end{equation}




\subsection{Softmin (Our Proposed Method)}

With a well-trained model, we expect the resulting per-pixel scores $s_{ij}$ to be lower for pixels that are incorrectly annotated or otherwise ambiguous looking. 
Even in perfectly labeled regions of an image, the $s_{ij}$ will inevitably vary due to natural statistical fluctuations (estimation error in model training).   Thus their mean, $s_{CIL}(x)$, may be undesirably sensitive to such $s_{ij}$ variations in correctly-labeled regions of an image (which presumably represent most of most images). This can be mitigated by focusing solely on the \emph{worst-labeled} region of an image, for instance the minimum of the $s_{ij}$ corresponds to one of the least confidently well-labeled pixels in the image. While robust to nuisance variation in the pixel-scores for correctly-labeled regions of the image, the minimum score entirely ignores most of the image, which is also undesirable. Presumably it is easier to determine certain images are mislabeled when a large region of many pixels all receive low scores (see Figure \ref{fig:cityscapes}).

Instead of taking the minimum of these scores, we can instead form a soft approximation of the minimum to remain robust to nuisance variation in high pixel-scores while still accounting for the scores across the entire image. Our \emph{softmin} scoring method interpolates between the mean and minimum score. It computes an overall label quality score for the image from per-pixel confidence values via a soft version of the minimum operator:
\begin{equation}
s_{SM}(x) = \sum_{i,j} s_{ij}  \cdot \frac{\exp\left(\frac{1-s_{ij}}{\tau}\right)}{\sum_{i,j} \exp\left(\frac{1-s_{ij}}{\tau}\right)}
\end{equation}
This is analogous to the popular softmax approximation of the argmax operator, where temperature parameter $\tau$ controls the sharpness of the minimum approximation above. 
Since the $s_{ij}$ are confidence values between 0 and 1, we use a smaller $\tau = 0.1$ throughout to appropriately tradeoff between emphasizing the smallest per-pixel scores (most indicative of annotation error locations within an image) while still accounting for the rest of the image. A similar setting was explored by \citet{eric} to diagnose mislabeling of text datasets used for entity recognition.

\subsection{Confident Learning Counts (CLC)}
The alternative \emph{Confident Learning Counts} label quality score extends the popular Confident Learning method for LED in classification \cite{northcutt2021confident} to segmentation settings. Treating each pixel as an independent instance in a classification task (ignoring which image it belongs to), we apply Confident Learning to infer a binary mask $\mathbf{b}$ estimating \emph{which} pixels in image $x$ are mislabeled. 
A resulting label quality score is naturally defined as the proportion of pixels in the image estimated to be mislabeled: 
\begin{equation}
    s_{CLC} = \frac{\sum_{i,j} b_{ij}}{h \cdot w}
\end{equation}
To reduce runtime and memory requirements of the method, we first downsample predicted/annotated segmentation masks 4 times before computing $s_{CLC}$. In downsampling, predicted class probabilities are mean pooled within each $4 \times 4$ grid, and the resulting annotated label for this grid is determined via majority vote over the $4 \times 4$ pixels. 

\subsection{IOU}
A standard evaluation measure to measure the similarity between segmentation predictions and labels is  the IoU metric (Intersection over Union, a.k.a.\ Jaccard Index) \cite{segmentanything, meta}. 
The IOU for an individual image can serve as a label quality score, calculated as:
\begin{equation}
s_{IOU}(x)= \frac{|\mathbf{P} \cap \mathbf{l}|}{|\mathbf{P} \cup \mathbf{l}|}
\end{equation}
where $\mathbf{P}$ represents the predicted segmentation mask and $\mathbf{l}$ represents the annotated mask. Sorting images based on $s_{IOU}$ corresponds to  traditional \emph{error analysis}, in which one inspects the highest loss images for suspicious anomalies, with loss computed with respect to model predictions based on a standard evaluation metric. 

\subsection{Connected Components (CoCo)}
\label{conn}
\citet{rottmann2022automated} suggest that LED on the pixel-level is less robust than scoring label quality at a connected component level. This approach exploits the spatial nature of segmentation data, where neighboring pixels often belong to the same class. In the \emph{Connected Components} approach, we produce a quality score for each component of an image, a spatially-contiguous region of pixels all annotated/predicted to share the same class. Subsequently we pool these component scores into an overall image score by taking their average over components and classes. Component scores are produced by producing a predicted class probability for the overall component and using it to evaluate the likelihood of the annotated label for the component. 

More specifically, we calculate an overall label quality score $s_{CoCo}(x)$ for image $x$ as follows:
\begin{enumerate}
    \item Form a set of connected components based on each unique combination of predicted and annotated segmentation masks $\mathbf{P}$ and $\mathbf{l}$.
    \item For each connected component $c$, use the mean of the predicted probabilities $p_{ijk}$ for the pixels within the component as a predicted class probability estimate for the entire component $p_c$.
    \item Compute a quality score for the component $s_c = p_c[k]$ where $k$ is the annotated class label for this component, using the same sort of label-likelihood we computed for each pixel $s_{i,j}$ in other approaches.
    \item Average the per-component scores $s_c$ across all components $c$ to obtain the overall label quality score $s_{CoCo}(x)$ for image $x$.
\end{enumerate}
To reduce runtime and memory requirements of the method, we first downsample predicted/annotated segmentation masks 4 times before computing $s_{CoCo}$. Here we follow the same downsampling procedure previously described.

\section{Experiments}

\subsection{Datasets}
\begin{figure}[t]
    \centering
    \begin{subfigure}{Drop (Car overlooked) - Given Mask for Unlabeled class}
        \centering
        \includegraphics[width=\linewidth]{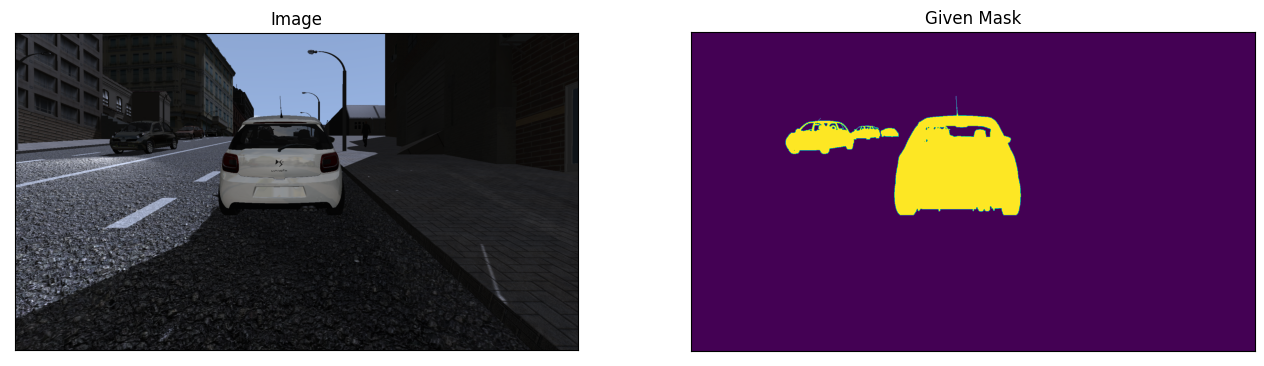}
    \end{subfigure}%
    \begin{subfigure}{Shift Error (in Sky class) - Given Mask for Sky class}
        \centering
        \includegraphics[width=\linewidth]{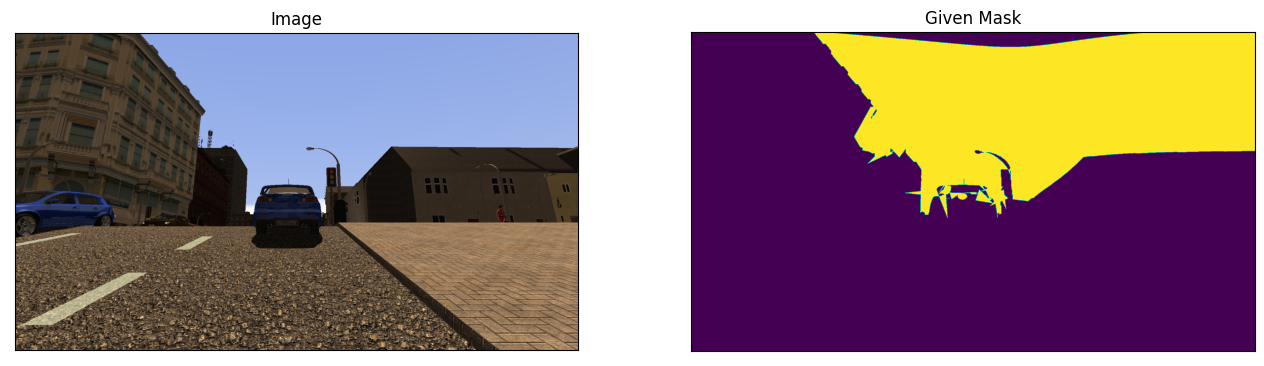}
    \end{subfigure}
    \begin{subfigure}{Swap (Building $\rightarrow$ Vegetation) - Given Mask for Vegetation}
        \centering
        \includegraphics[width=\linewidth]{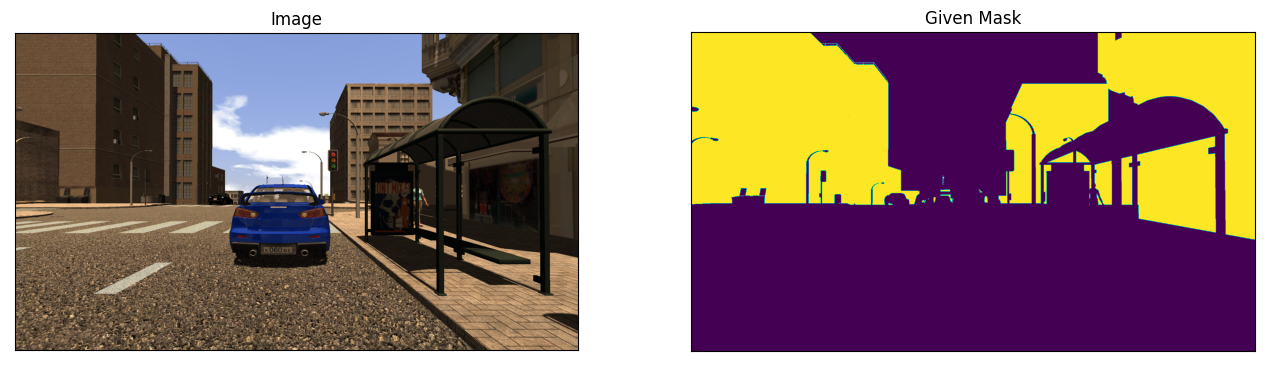}
    \end{subfigure}
    \vspace*{-1.3em}
    \caption{Examples of the three types of annotation errors in our SYNTHIA datasets. In the depicted \emph{Drop} error, the car was overlooked by annotators and its pixels are left as part of the Unlabeled class. In the depicted \emph{Shift} error, the mask for the Sky class  was poorly drawn and incorrectly covers part of the car. In the depicted \emph{Swap} error, the buildings have been mislabeled as the  Vegetation class instead of the Building class. All three of these images are amongst the  top-10 images with lowest softmin label quality score in the dataset, and thus automatically prioritized for review.}
    \label{fig:main}
\end{figure}

Most real-world segmentation data are full of annotation errors \cite{vuadineanu2022analysis}, but we must benchmark our label quality scores on a dataset where we can be sure of the underlying true segmentation masks. Here we use the SYNTHIA dataset \cite{synthia}, a simulated vehicular dataset similar to Cityscapes  \cite{cityscapes}, but with images generated via graphics engine. To study mislabeling that reflects common issues encountered in real-world segmentation data, we introduce three types of errors into the given labels of our dataset (depicted in Figure \ref{fig:main}):
\begin{enumerate}
\item \textbf{Drop}: Randomly eliminate the label of a selected class, mapping those pixels to the ``unlabeled'' category. The \emph{Drop} error mimics situations in which annotators forget to label portions of an image, overlooking certain objects, or forgetting the class exists in the set of choices.
\item \textbf{Swap}: Randomly interchange the labels of two selected classes across all pixels of the chosen image. The \emph{Swap} error mimics situations where annotators possess low certainty about which class to select or accidentally select the wrong one, perhaps not realizing a better class label exists in the set of categories.
\item \textbf{Shift}: Inspired by \citet{perturbations}, this error introduces variability in the shape of a mask for a chosen class, specifically along its edges. Using the OpenCV library \cite{opencv_library}, we introduce random morphological variation in the segmentation masks, representing annotators who  sloppily draw the labels. 
\end{enumerate}

Our benchmark has three datasets, each containing one of these types of labeling errors. The set of images is the same in each dataset, but differing  between datasets is: which images are mislabeled, the types of label errors, and the proportion of mislabeled images. In our first, second, and third dataset: 20\%, 30\%, and 20\%  of the images respectively have a \emph{Drop}, \emph{Swap}, or \emph{Shift} error. These settings allow us to characterize label error detection performance separately across the three error types, facilitating fine-grained study of our proposed methods under diverse conditions observed in real-world segmentation data \cite{vuadineanu2022analysis}.

In each dataset, label errors are randomly introduced before establishing  training and validation splits. Each training and each validation dataset has 1,112 images. In our benchmark, we only produce predictions and label quality scores for images in the validation set -- these are the only images considered in our evaluations. Note that one could instead employ cross-validation to score label quality of every image, but the benchmark conclusions would unlikely change.

\subsection{Models}

To study how well LED methods generalize across different segmentation models, we apply each label quality scoring technique with two different models:
\vspace*{-0.2em}
\begin{enumerate}
    \item DeepLabV3+ \cite{deeplab}: enhances the DeepLabV3 model with a decoder module to further refine segmentation results, particularly along object boundaries. This model has shown excellent performance in semantic segmentation tasks and has been widely used in various applications.
    \item FPN \cite{lin2017feature} The Feature Pyramid Network is a strong feature extractor for multi-scale object detection. The FPN model handles scale variation by using a top-down pathway and lateral connections to combine low-resolution, semantically strong features with high-resolution, semantically weak features.
\end{enumerate}
Both models are implemented using the PyTorch backbone provided by \citet{Iakubovskii} (with the ``se\_resnext50\_32x4d'' encoder and ``imagenet'' pretrained weights). The final activation function for both models is ``softmax2d''. 
All three models are fit to the noisily labeled training data each with their respective error type and used to produce out-of-sample predictions for a noisy validation set over which we evaluate LED performance.
Our experiments reflect a common setup in practical segmentation applications, in order to understand how effectively label errors can be detected under this setup.

\subsection{Evaluation}

Detecting the mislabeled images within a large dataset is an information retrieval task, and thus we evaluate performance with standard precision/recall metrics. 
In particular, we evaluate how well our label quality scores are able to \emph{rank} truly mislabeled images above those which are correctly labeled via: the Area Under the Receiver Operating Characteristic Curve (\textbf{AUROC}), the Area Under the Precision-Recall Curve (\textbf{AUPRC}), and \textbf{Lift @ T}.  Lift measures how many times more prevalent labels errors are within the top-$T$ scoring images vs.\ all images (and is monotonically related to \textbf{Precision @ T}. We consider $T = 100$ or setting it equal to the true number of mislabeled images in each dataset.

Evaluation via Lift assesses scores' precision, while AUROC and AUPRC assess 
both precision and recall. In settings where true positives are relatively rare (as our mislabeled images are here), \citet{saito2015precision} argue AUPRC is more informative than AUROC.

\section{Results}
Tables \ref{AUROC}-\ref{lift100} report results from a comprehensive evaluation of all aforementioned label quality methods using both models on our variants of the SYNTHIA dataset. 
The results reveal that our \emph{Softmin} approach consistently delivers the most (or second-most in certain settings) effective detection of labeling errors, irrespective of the error type or the specific model in use. 
Our proposed method consistently demonstrates superior performance compared to existing strategies that rely on variations of the IoU metric \cite{rottmann2022automated}. While the \emph{Connected Components} approach does fare slightly better at detecting \emph{Shift} errors according to some metrics,  \emph{Connected Components} label quality scores do not work well for detecting  \emph{Drop} errors.
We find that detecting \emph{Swap} errors is significantly easier than other two types of error, with \emph{Shift} errors being hardest to detect. Visually, we also found this to be the case when manually inspecting our SYNTHIA images for errors.


\begin{table*}[t!]
\caption{AUROC achieved by various label quality scores used with two types of models for detecting three types of annotation errors.}
\label{AUROC}
\vskip 0.15in
\begin{center}
\begin{sc}
\begin{tabular}{lcccccc}
\toprule
Method                          & \multicolumn{2}{c}{Drop}     & \multicolumn{2}{c}{Swap}       & \multicolumn{2}{c}{Shift}      \\
\midrule
Model                           & DeepLabV3+ & FPN & DeepLabV3+ & FPN & DeepLabV3+ & FPN \\
\midrule
Correctly Classified Pixels                         & 0.915                         & 0.916                   & \textbf{ 1.000  }                       & \textbf{ 1.000 }                  & 0.833                         & 0.798                   \\
 Thresholded CCP              & 0.869                         & 0.869                   & 0.993                         & 0.993                   & 0.852                         & 0.818                   \\
Confidence In Label         & 0.915                         & 0.916                   & \textbf{ 1.000 }                         & \textbf{ 1.000 }                   & 0.832                         & 0.797                   \\
Confident Learning Counts & 0.904                         & 0.905                   & \textbf{ 0.999 }                       & \textbf{ 0.999 }                   & 0.807                         & 0.778       \\  
IoU                             & 0.921                         & 0.901                   & 0.880                         & 0.880                   & 0.713                         & 0.649                   \\
Connected Components               & 0.880                         & 0.888                   & 0.984                         & 0.982                   & 0.783                         & 0.819                   \\
Softmin            & \textbf{ 0.951 }                        & \textbf{ 0.947  }                 & \textbf{ 0.998 }                        &  \textbf{ 0.998 }                  & \textbf{ 0.863 }                        & \textbf{ 0.828 }                  \\
\bottomrule
\end{tabular}
\end{sc}
\end{center}
\vskip -0.1in
\end{table*}

\subsection{Annotation errors in the CityScapes Dataset}

\begin{figure}[tb]
    \centering
    \begin{subfigure}{Road overlooked - Given Mask for Unlabeled class}
        \centering
        \includegraphics[width=\linewidth]{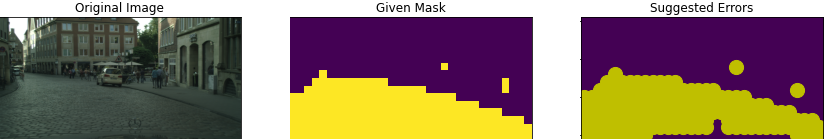}
    \end{subfigure}%
    \begin{subfigure}{Sky overlooked - Given Mask for Unlabeled class}
        \centering
        \includegraphics[width=\linewidth]{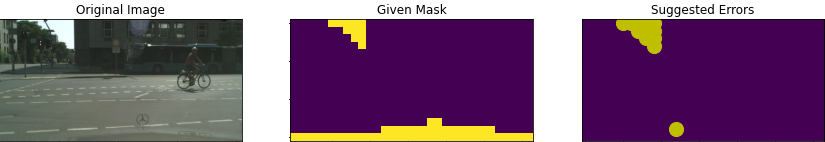}
    \end{subfigure}
    \begin{subfigure}{Car overlooked - Given Mask for Unlabeled class}
        \centering
        \includegraphics[width=\linewidth]{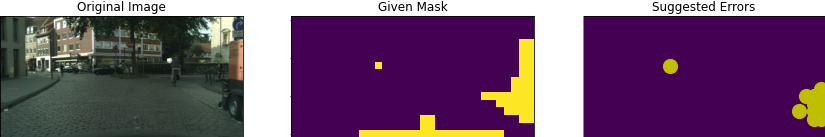}
    \end{subfigure}
    \begin{subfigure}{Sidewalk overlooked - Given Mask for Unlabeled class}
        \centering
        \includegraphics[width=\linewidth]{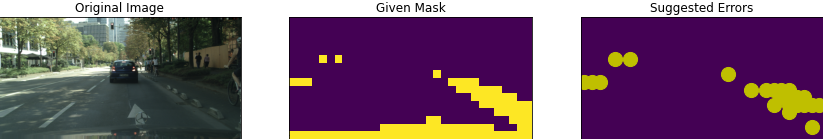}
    \end{subfigure}
    \vspace*{-1.5em}
    \caption{Images which received some of the lowest softmin label quality scores in the Cityscapes dataset. 
    In different images, the \textit{Road, Sky, Car,} and \textit{Sidewalk} were overlooked by annotators -- pixels which should have been annotated with these classes were mistakenly left as the  \textit{Unlabeled} class. 
    The Suggested Errors on right highlight the pixels with low $s_{ij}$ values in each image.}
    \label{fig:cityscapes}
    \vspace*{-0.5em}
\end{figure}

Finally, we apply our methodology to audit for mislabeling in the popular CityScapes segmentation dataset, here studying its \emph{fine}-annotations  \cite{cityscapes}. 
We fit a DeepLabV3+ model to this data and use the model's out-of-sample predictions with our softmin score to assess each image's label quality. 
Figure \ref{fig:cityscapes} shows top \emph{naturally-occurring} label errors discovered by this approach, and many more of such overlooked errors were discovered. The full results for CityScapes are provided in the previously linked benchmarks GitHub repository. 

\section{Discussion}

Our study presents a comprehensive evaluation of eight different label quality scoring methods for semantic segmentation datasets. Based on a soft-minimum of the model-estimated likelihoods of each pixel's annotated class, the \emph{Softmin} score is particularly effective for detecting mislabeled images with high precision and recall, regardless what types of annotation errors lurk in the data. This highlights the value of focusing on potentially severely mislabeled regions of an image when estimating which images have imperfect annotations. 
Our findings align with insights from \citet{eric}, who found taking a minimum over per-token scores to be an effective sentence label quality score for entity recognition (text) data.

The \emph{Softmin} score is easy to apply with any trained segmentation model, and thus will remain applicable as improved segmentation architectures and training procedures are invented. After using this method to detect annotation errors, they can be fixed or some of the bad data omitted from the training set, and subsequently a more reliable copy of this same model trained without any change in the modeling code.  
As the accuracy of models increases via new innovations, the LED performance of our label quality scores will improve, thus enabling \emph{even} more accurate versions of these models to be produced from higher integrity data.


\begin{table*}[t!]
\caption{AUPRC achieved by various label quality scores used with two types of models for detecting three types of annotation errors.}
\label{AUPRC}
\vskip 0.15in
\begin{center}
\begin{sc}
\begin{tabular}{lcccccc}
\toprule
Error Type                          & \multicolumn{2}{c}{Drop}     & \multicolumn{2}{c}{Swap}       & \multicolumn{2}{c}{Shift}      \\
\midrule
Model                           & DeepLabV3+ & FPN & DeepLabV3+ & FPN & DeepLabV3+ & FPN \\
\midrule
Correctly Classified Pixels                        & 0.814                                  & 0.814                            & \textbf{ 0.999 }                                  & \textbf{ 0.999 }                           & 0.474                                  & 0.399                            \\
Thresholded CCP              & 0.684                                  & 0.681                            & 0.983                                  & 0.983                            & 0.526                                  & 0.440                            \\
Confidence In Label         & 0.813                                  & 0.814                            & \textbf{ 0.999  }                                & \textbf{ 0.999 }                           & 0.472                                  & 0.398                            \\
Confident Learning Counts & 0.795                                  & 0.796                            & \textbf{ 0.999 }                                 & \textbf{ 0.999 }                           & 0.429                                  & 0.375     \\                      
IoU                            & 0.808                                  & 0.749                            & 0.754                                  & 0.759                            & 0.440                                  & 0.320                            \\
Connected Components               & 0.654                                  & 0.675                            & 0.971                                  & 0.965                            & 0.519                                  & \textbf{ 0.537 }                            \\
Softmin            & \textbf{ 0.888 }                                 & \textbf{ 0.875 }                           & \textbf{ 0.996 }                                  & \textbf{ 0.996 }                            & \textbf{ 0.545  }                                & 0.461                            \\
\bottomrule
\end{tabular}
\end{sc}
\end{center}
\vskip -0.1in
\end{table*}

\begin{table*}[h!]
\caption{Lift @ $T$ achieved by various label quality scores used with two types of models for detecting three types of annotation errors. Here $T$ is set equal to the true number of mislabeled images in each dataset.}
\label{lifterrors}
\vskip 0.15in
\begin{center}
\begin{sc}
\begin{tabular}{lcccccc}
\toprule
Error Type                          & \multicolumn{2}{c}{Drop}     & \multicolumn{2}{c}{Swap}       & \multicolumn{2}{c}{Shift}      \\
\midrule
Model                           & DeepLabV3+ & FPN & DeepLabV3+ & FPN & DeepLabV3+ & FPN \\
\midrule
Correctly Classified Pixels                        & 3.637                                  & 3.662                            & \textbf{ 3.369 }                                 & \textbf{3.369    }                        & 2.391                                  & 2.254                            \\
Thresholded CCP               & 3.216                                  & 3.117                            & 3.167                                  & 3.157                            & 2.709                                  & 2.345                            \\
Confidence In Label         & 3.637                                  & 3.662                            & \textbf{ 3.369 }                                  & \textbf{ 3.369 }                            & 2.413                                  & 2.254                            \\
Confident Learning Counts & 3.588                                  & 3.588                            & \textbf{ 3.369  }                                & 3.358                            & 2.163                                  & 2.072               \\ 
IoU                             & 3.983                                  & 3.464                            & 2.341                                  & 2.320                            & 2.254                                  & 1.935                            \\
Connected Components               & 3.142                                  & 3.167                            & 3.114                                  & 3.061                            & 2.550                                  & \textbf{ 2.618 }                           \\           
Softmin            & \textbf{ 4.231 }                                  & \textbf{ 4.107 }                            & 3.305                                  & 3.305                            & \textbf{ 2.755 }                                  &  2.550                             \\
\bottomrule
\end{tabular}
\end{sc}
\end{center}
\vskip -0.1in
\end{table*}

\clearpage 
\begin{table*}[h!]
\caption{Lift @100 achieved by various label quality scores used with two types of models for detecting three types of annotation errors.}
\label{lift100}
\vskip 0.15in
\begin{center}
\begin{sc}
\begin{tabular}{lcccccc}
\toprule
Error Type                          & \multicolumn{2}{c}{Drop}     & \multicolumn{2}{c}{Swap}       & \multicolumn{2}{c}{Shift}      \\
\midrule
Model                           & DeepLabV3+ & FPN & DeepLabV3+ & FPN & DeepLabV3+ & FPN \\
\midrule
Correctly Classified Pixels                        & 5.088                                  & 5.088                            & \textbf{ 3.432 }                                  & \textbf{ 3.432 }                            & 2.315                                  & 2.113                            \\
Thresholded CCP               & 4.196                                  & 4.144                            & \textbf{ 3.432 }                                 & \textbf{ 3.432 }                            & 2.767                                  & 2.365                            \\
Confidence In Label        & 5.088                                  & 5.088                            & \textbf{ 3.432 }                                 & \textbf{ 3.432 }                           & 2.315                                  & 2.113                            \\
Confident Learning Counts & 4.983                                  & 4.983                            & \textbf{ 3.432  }                                & \textbf{ 3.432 }                           & 2.164                                  & 2.013      \\                     
IoU                             & 4.773                                  & 4.511                            & 3.020                                  & 3.055                            & 3.120                                  & 2.164                            \\
Connected Components              & 3.882                                  & 3.934                            & \textbf{ 3.432 }                                  & \textbf{ 3.432 }                           &  \textbf{3.170}                                  & \textbf{ 3.271 }                           \\
Softmin            & \textbf{ 5.140 }                                  & \textbf{ 5.193 }                            & \textbf{ 3.432  }                                & \textbf{ 3.432 }                            & 3.019                                  & 2.315                            \\
\bottomrule
\end{tabular}
\end{sc}
\end{center}
\vskip -0.1in
\end{table*}

\FloatBarrier
\balance{}
\bibliography{semseg}
\bibliographystyle{icml2023}

\end{document}